\documentclass[conference]{IEEEtran}
\IEEEoverridecommandlockouts
\usepackage{cite}
\usepackage{amsmath,amssymb,amsfonts}
\usepackage{algorithmic}
\usepackage{graphicx}
\usepackage{textcomp}
\usepackage{xcolor}
\usepackage{cleveref}
\usepackage[utf8]{inputenc}
\usepackage[normalem]{ulem}
\usepackage{url}

\usepackage{xcolor}

\def\BibTeX{{\rm B\kern-.05em{\sc i\kern-.025em b}\kern-.08em
    T\kern-.1667em\lower.7ex\hbox{E}\kern-.125emX}}
\begin{document}

\title{Bayesian Autoencoders for Drift Detection in Industrial Environments}

\author{\IEEEauthorblockN{Bang Xiang Yong}
\IEEEauthorblockA{\textit{Institute for Manufacturing} \\
\textit{University of Cambridge}\\
Cambridge, United Kingdom \\
bxy20@cam.ac.uk}
\and
\IEEEauthorblockN{Yasmin Fathy}
\IEEEauthorblockA{\textit{Institute for Manufacturing} \\
\textit{University of Cambridge}\\
Cambridge, United Kingdom \\
yafa2@cam.ac.uk}
\and
\IEEEauthorblockN{Alexandra Brintrup}
\IEEEauthorblockA{\textit{Institute for Manufacturing} \\
\textit{University of Cambridge}\\
Cambridge, United Kingdom \\
ab702@cam.ac.uk}}
\bibliographystyle{IEEEtran}
\maketitle

\begin{abstract}

Autoencoders are unsupervised models which have been used for detecting anomalies in multi-sensor environments. A typical use includes training a predictive model with data from sensors operating under normal conditions and using the model to detect anomalies. Anomalies can come either from real changes in the environment (real drift) or from faulty sensory devices (virtual drift); however, the use of Autoencoders to distinguish between different anomalies has not yet been considered. To this end, we first propose the development of Bayesian Autoencoders to quantify epistemic and aleatoric uncertainties. We then test the Bayesian Autoencoder using a real-world industrial dataset for hydraulic condition monitoring. The system is injected with noise and drifts, and we have found the epistemic uncertainty to be less sensitive to sensor perturbations as compared to the reconstruction loss. By observing the reconstructed signals with the uncertainties, we gain interpretable insights, and these uncertainties offer a potential avenue for distinguishing real and virtual drifts.

\end{abstract}

\begin{IEEEkeywords}
uncertainty, Bayesian autoencoder, deep learning, sensors
\end{IEEEkeywords}

\section{Introduction}
In smart factories, machine learning algorithms are increasingly used to extract values from multi-sensor data that monitor manufacturing processes whose characteristics are often complex and non-linear. In typical predictive models for manufacturing, the trust and safety in predictive models should be improved. As such, it is crucial to quantify and explain the confidence of the outcomes for the predictive models. An emerging area of research is the uncertainty quantification of deep learning models \cite{ghahramani2015probabilistic}. Primarily, there are two types of uncertainties, epistemic and aleatoric - the former is the uncertainty in the model parameters due to limited data availability, while the latter arises from the noise in data \cite{kendall2017uncertainties}. Nonetheless, to this end, the uncertainty of deep learning models remains understudied in most of realistic Industry 4.0 applications. 

Due to the dynamic and ad-hoc environment of factories, the collected data by multiple sensors are often non-stationary \cite{yong2019multi}. In condition monitoring and quality prediction, machine learning (ML) methods rely on quantifying and detecting real changes in the environment and object of interest (real drift). As sensors degrade over time with increasing noise and drift level, ML methods which rely on the measurements of these sensors are affected (virtual drift).

In the occurrence of drifts (real and virtual), the noise level of the underlying distribution may change. Hence, the model must capture the change, which is termed heteroscedastic aleatoric uncertainty. In contrast, in a model where the estimated noise level is assumed constant, it is called homoscedastic aleatoric uncertainty \cite{kendall2017uncertainties}.

Unsupervised deep learning models such as autoencoders have been shown to perform well for detecting real drifts in quality predictive \cite{wang2019generative} and prognosis applications \cite{martinez2019visually}. The models can also detect virtual drifts that are caused by faults in the sensors \cite{yang2018convolutional}. It's crucial to distinguish between real and virtual drifts such that operators can take appropriate mitigation actions depending on the source of anomalies.

This study is the first to shed light on the behaviour of quantified epistemic and aleatoric uncertainties in autoencoders for unsupervised learning within the context of real and virtual drift in sensor data. 

The paper is structured as follows. Section~\ref{autoencoder} provides the required background and our proposed approach. The performance evaluation including dataset, reproducibility of experimental results and evaluation are included in Section~\ref{experiments} and Section~\ref{results}. We conclude the paper and explain the future directions of our research in Section~\ref{conclusion}.

\section{Bayesian Autoencoder}
\label{autoencoder}

%
The general structure of an autoencoder maps a given set of unlabelled training data; $X = {\{x_1,x_2,x_3,...x_N\}}$, $x_{i} \in \rm I\!R^{D}$ into an output $\hat{x}$ (i.e reconstructed signal), through a latent representation, $h$~\cite{goodfellow2016deep}. Structurally, every autoencoder consists of two parts: an encoder $f$ for mapping original data $x$ into $h$ (i.e. $h= f(x)$) and a decoder $g$ for mapping $h$ to a reconstructed signal of the input $\hat{x}$ (i.e. $\hat{x} = g(h) = g(f(x)$). 

Based on Bayes rule,
\begin{equation}\label{eq_posterior}
    p(\theta|X) = \frac{p(X|\theta)\  p(\theta)}{p(X)} \\
\end{equation}

where $p(X|\theta)$ is the data likelihood which can be modelled as a diagonal  Gaussian distribution with i.i.d assumption and the likelihood mean is the Bayesian Autoencoder's output. $p(\theta)$ is the prior distribution of the Bayesian Autoencoder's parameters. For simplicity, one can assume a diagonal Gaussian prior which corresponds to an L2 regularisation. Since \cref{eq_posterior} is analytically intractable for a deep neural network which is highly non-linear and consists of a large number of parameters compared to classical statistical models. To this end, various approximate methods were developed such as Markov Chain Monte Carlo (MCMC) \cite{chen2014stochastic}, variational inference \cite{blundell2015weight}, MC Dropout \cite{gal2016dropout} and ensembling \cite{pearce2018uncertainty} to sample from the posterior distribution. Although these methods have been explored within supervised neural networks, to the best of our knowledge, they have not been extensively applied on autoencoders which are unsupervised models. Within these methods, the marginal distribution, $p(X)$ (or evidence) is often assumed as a constant and ignored. 

In this paper, we employ a sampling method, `anchored ensembling'\cite{pearce2018uncertainty} for approximating the posterior distribution while training the autoencoders. In anchored ensembling, posteriors are approximated by Bayesian inference under the family of methods called randomised maximum a posteriori (MAP) sampling, where model parameters are regularised by values drawn from a distribution (so-called anchor distribution), which can be set equal to the prior.

Assume our ensemble consists of M independent autoencoders and each j-th autoencoder contains a set of parameters, $\theta_j$ where $j\in{\{1,2,3...M\}}$. The autoencoders are trained by minimising the loss function, which is the negative sum of log-likelihood (based on i.i.d assumption) and log prior where both are assumed to be Gaussian. The loss due to the likelihood is: 

\begin{equation} \label{eq_lik_loss}
\mathcal{L}(X, \hat{X}) = \frac{1}{N} \sum^{N}_{i=1}{\frac{1}{2\sigma_i^2}}||x_i-\hat{x_i}||^2 + \frac{1}{2}\log{\sigma_i^2}
\end{equation}

where $\sigma_i^2$ is the variance of the data point, which is also known as aleatoric uncertainty for regression tasks. Note that a typical autoencoder minimises the reconstruction loss, $||x_i-\hat{x_i}||^2$ which corresponds to a diagonal Gaussian likelihood with a fixed variance of 1. Instead of a fixed variance, by `learning` the variance term as an output of the autoencoder, the model can estimate the noise level for every data point $x_i$. Following the method proposed by \cite{kendall2017uncertainties} to compute heteroscedastic aleatoric uncertainty for regression tasks, an extra layer is added to the final layer of autoencoder with dimensions equal to the size of the inputs, to predict the log variance, $\log{\sigma_i^2}$ corresponding to each data point $x_i$. 

In anchored ensembling for approximating the posterior distribution, the `anchored weights` for each autoencoder are unique and sampled during initialisation from a prior distribution $\theta_{anc,j} \sim N(\mu_{anc,j},\sigma_{anc,j}^2)$ and remain fixed throughout the training procedure. To scale the regulariser term arising from the prior, $\lambda$ is set as a hyperparameter. The loss due to prior is:

\begin{equation} \label{eq_prior_loss}
\mathcal{L}(\theta_j) = \frac{\lambda}{N} \sum^{N}_{i=1}||\theta_j-\theta_{anc,j}||^2
\end{equation}


With \cref{eq_lik_loss} and \cref{eq_prior_loss}, the resulting loss function to be minimised is:

\begin{equation} \label{eq_final_loss}
\mathcal{L}(X, \hat{X}, \theta_j) = \mathcal{L}(X, \hat{X}) + \mathcal{L}(\theta_j)
\end{equation}


For model prediction, the predictive posterior distribution of an unseen test input $X^*$, is calculated by integrating over all possible $\theta$: 

\begin{equation}\label{eq_predictive_full}
p(X^*|X) = \int{p(X^*|\theta,X)\ p(\theta|X)\ d\theta}
\end{equation}

Although ~\cref{eq_predictive_full} is intractable, we can estimate it with the samples of $p(\theta|X)$ which we obtained by training the ensemble: 

\begin{equation}\label{eq_predictive_estimate}
\hat{p}(X^*|X) = \sum_{j=1}^{M}{p(X^*|\theta_j,X)}
\end{equation}



To compute epistemic uncertainty on a new single data point, $x^*$, the variance of reconstructed signals from the ensemble is computed: 

\begin{equation}
Var(\hat{x}^*) = \frac{\sum_{j=1}^M{(\hat{x}^*_j-\bar{x})^2}}{M} 
\end{equation}
where M is the number of ensembled autoencoders, $\bar{x}$ is the mean of reconstructed signals $\hat{x}^*_j$.

In addition to the reconstructed signals, the Bayesian Autoencoder also outputs the log variance of data, $\log{\sigma_i}$, by which we can recover the heteroscedastic aleatoric uncertainty, $\sigma_i^2$ with the exponential function.

\section{Experimental Evaluation}
\label{experiments}
This section explains the real-world dataset used in our evaluation, the reproducibility of our results and the evaluation criteria of our proposed method. 

\subsection{Dataset}
We have tested our proposed approach on a publicly available dataset for condition monitoring of hydraulic system \cite{helwig2015}. The dataset is obtained from a hydraulic test rig which permits safe and non-destructive changes to the states of various components (cooler, valve, pump and accumulator) to emulate faults and degradation. Redundant sensors are equipped on the test system on multiple locations to measure pressure, flow, temperature, power and vibration. There is a total of 17 sensors and various sampling frequencies of 1Hz, 10Hz and 100Hz. In the dataset, there is a total of 2205 cycles, and each working cycle of the hydraulic system lasts for 60 seconds. The methods developed in our study are not limited to condition monitoring and can be applied to other Industry 4.0 use cases. 

\subsection{Data processing}
Due to the inconsistent sampling frequencies, we resample the data to 1Hz. As such, this results in 60 (time points) * 17 (sensors) for each cycle. The features are then normalised using a standard scaler for each sensor with careful implementations to prevent train-test bias. We do not compute specific features from the data but instead we feed the resampled and rescaled raw signals to the Bayesian Autoencoder. By doing so, we are able to visualise and gain insights into the full reconstructed signals as predicted by the deep model.

\subsection{Experiment setup}
We set the number of hidden nodes and layers of the Bayesian Autoencoder to 1020-500-250-3-250-500-1020 with 10 samples in the ensemble. The Bayesian Autoencoder is trained and tested with 70\% and 30\%, respectively of the sensor data where the cooler condition is known to be healthy. Then, for the case of real drift, we test the model on data which the cooler condition has degraded to 20\% and 3\% (near failure) efficiency. To simulate virtual drifts scenarios, we create two datasets from the `healthy` test set and artificially inject a range of noise from 5-25\% (i.e. injected noise of uniform distribution) and constant sensor drift of 5-25\% of the mean in each one of the sensors (i.e. injected drift). 

To ensure the reproducibility of our results, we have
made the code of our implementation available and have also provided details of a configurable experimental set-up at~\url{https://github.com/bangxiangyong/bae-drift-detection-zema-hydraulic}.

\section{RESULTS AND DISCUSSION}

\label{results}
\begin{figure}[htbp]
\centerline{\includegraphics[scale=0.6]{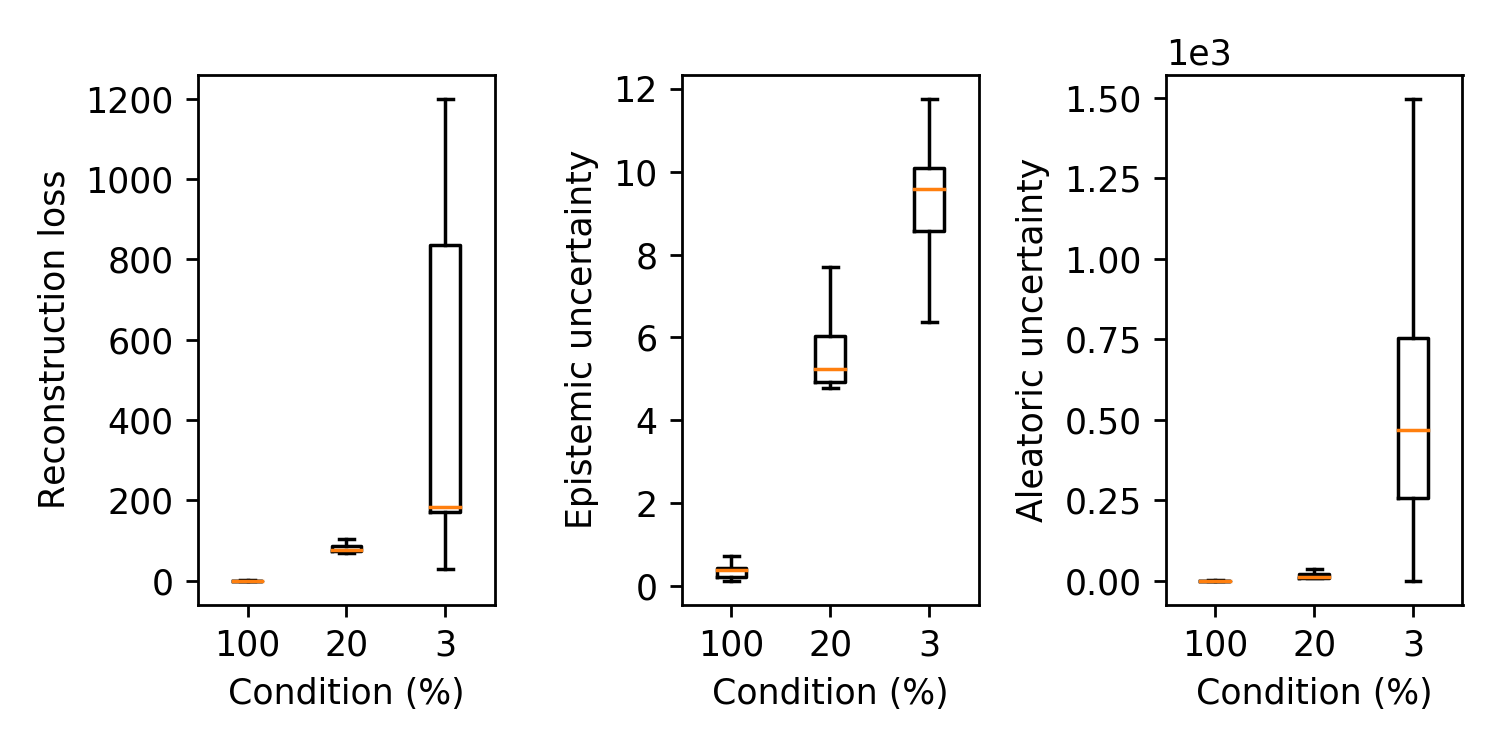}}
\caption{a) Reconstruction loss, b) Epistemic uncertainty, c) Aleatoric uncertainty under real drift of degrading cooler condition}
\label{fig-real-drift}
\end{figure}

\begin{figure}[htbp]
\centerline{\includegraphics[scale=0.6]{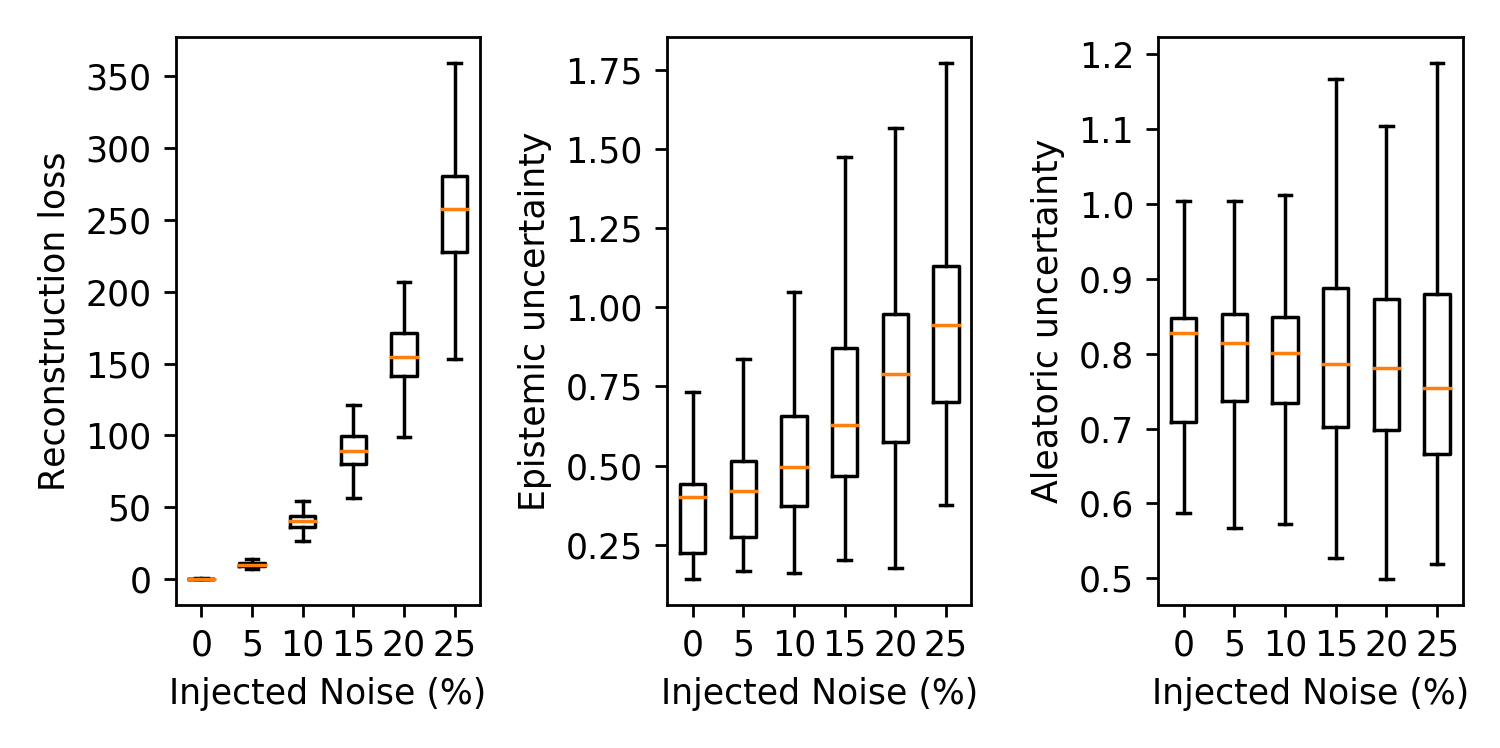}}
\caption{a) Reconstruction loss, b) Epistemic uncertainty, c) Aleatoric uncertainty under virtual drift of increasing noise in a pressure sensor}
\label{fig-injected-noise}
\end{figure}

\begin{figure}[htbp]
\centerline{\includegraphics[scale=0.6]{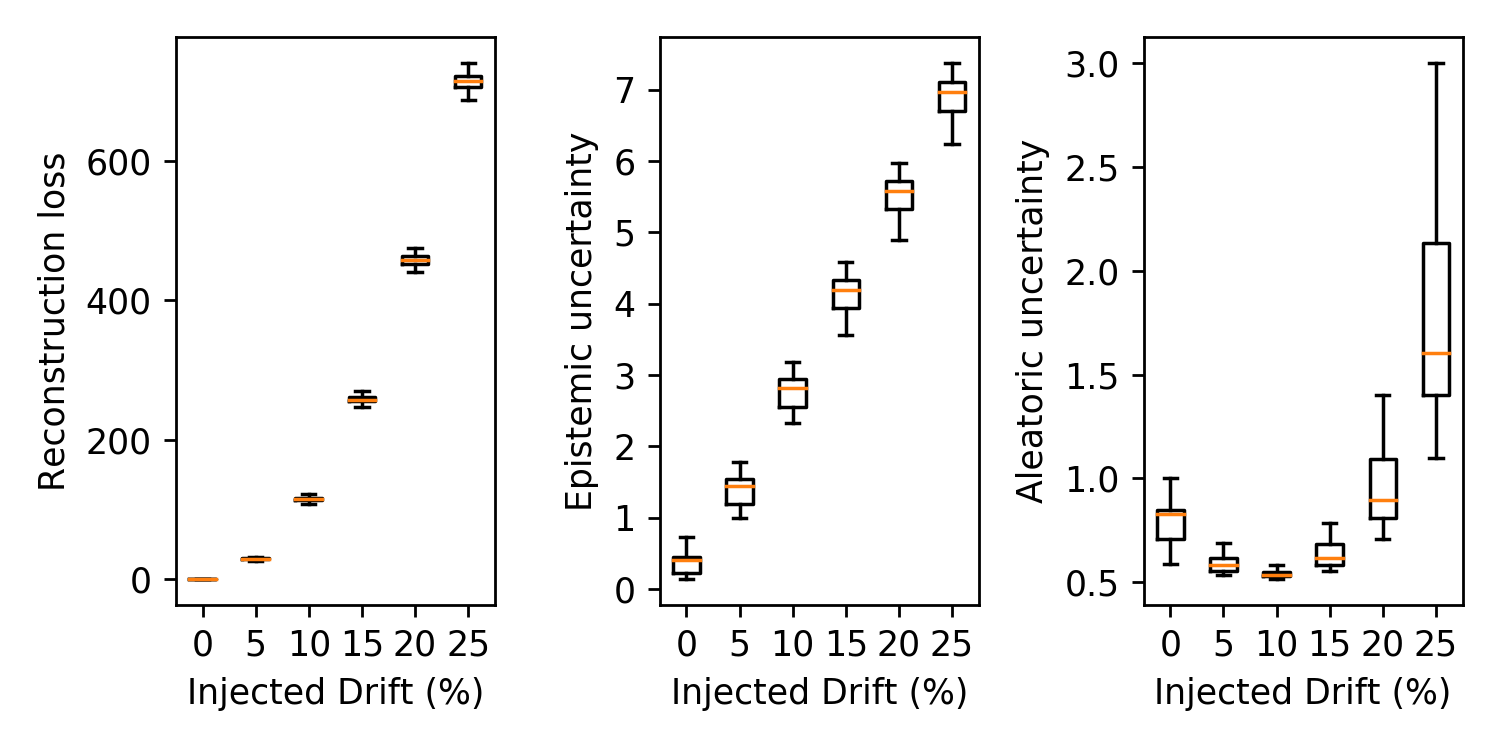}}
\caption{a) Reconstruction loss, b) Epistemic uncertainty, c) Aleatoric uncertainty under virtual drift of increasing drift in a pressure sensor}
\label{fig-injected-drift}
\end{figure}

We have conducted three sets of experiments; 1) real drift, 2) injected noise, and 3) injected drift. The reconstruction loss, epistemic and aleatoric uncertainties for these experiments are summarised in Fig.~\ref{fig-real-drift}, Fig.~\ref{fig-injected-noise} and Fig.~\ref{fig-injected-drift} respectively. Although we show the results for only one of the pressure sensors, denoted PS1, we have extended the experiment to every sensor and found similar results. One limitation of our analysis is we do not explore virtual drifts on combination of sensors.
In general, we note that the mean of reconstruction loss, epistemic and aleatoric uncertainties increase in both cases of real and virtual drift conditions. 

The reconstruction loss for the cooler condition of \%3 has a longer tail which overlaps with the less faulty conditions (Fig.~\ref{fig-real-drift}a). In practice, when using it for anomaly detection, this may lead to false positives. Additionally, the reconstruction loss and aleatoric uncertainty increase exponentially with the degrading condition of cooler, whereas epistemic uncertainty increases linearly in the same scenarios (Fig.~\ref{fig-real-drift}). 

Moreover, the epistemic uncertainty is generally less affected by noise in the sensor than the reconstruction loss (Fig.~\ref{fig-injected-noise}). Unexpectedly, however, in the advent of increasing sensor noise, the aleatoric uncertainty does not increase, as shown in Fig.~\ref{fig-injected-noise}c. Intuitively, we expect the estimated variance to be proportional to the level of sensor noise. In contrast, we note that the aleatoric uncertainty increases dramatically for the degrading cooler condition since multiple sensors are affected simultaneously. Therefore, for comparing these two situations, an exploration step is to investigate the effects of perturbations applied on a combination of sensors. With that, we can develop a feature importance ranking based on the sensitivity of the model's uncertainties. 

In the case of injected sensor drift (Fig.~\ref{fig-injected-drift}), the reconstruction loss increases exponentially, whereas the epistemic uncertainty increases almost linearly. In contrast, aleatoric uncertainty shows a convex behaviour. Unfortunately, since the aleatoric uncertainty is computed using a black-box model, we do not have an intuitive explanation for it \cite{venkatesh2019heteroscedastic}. 

\begin{figure}[htbp]
\centerline{\includegraphics[scale=0.70]{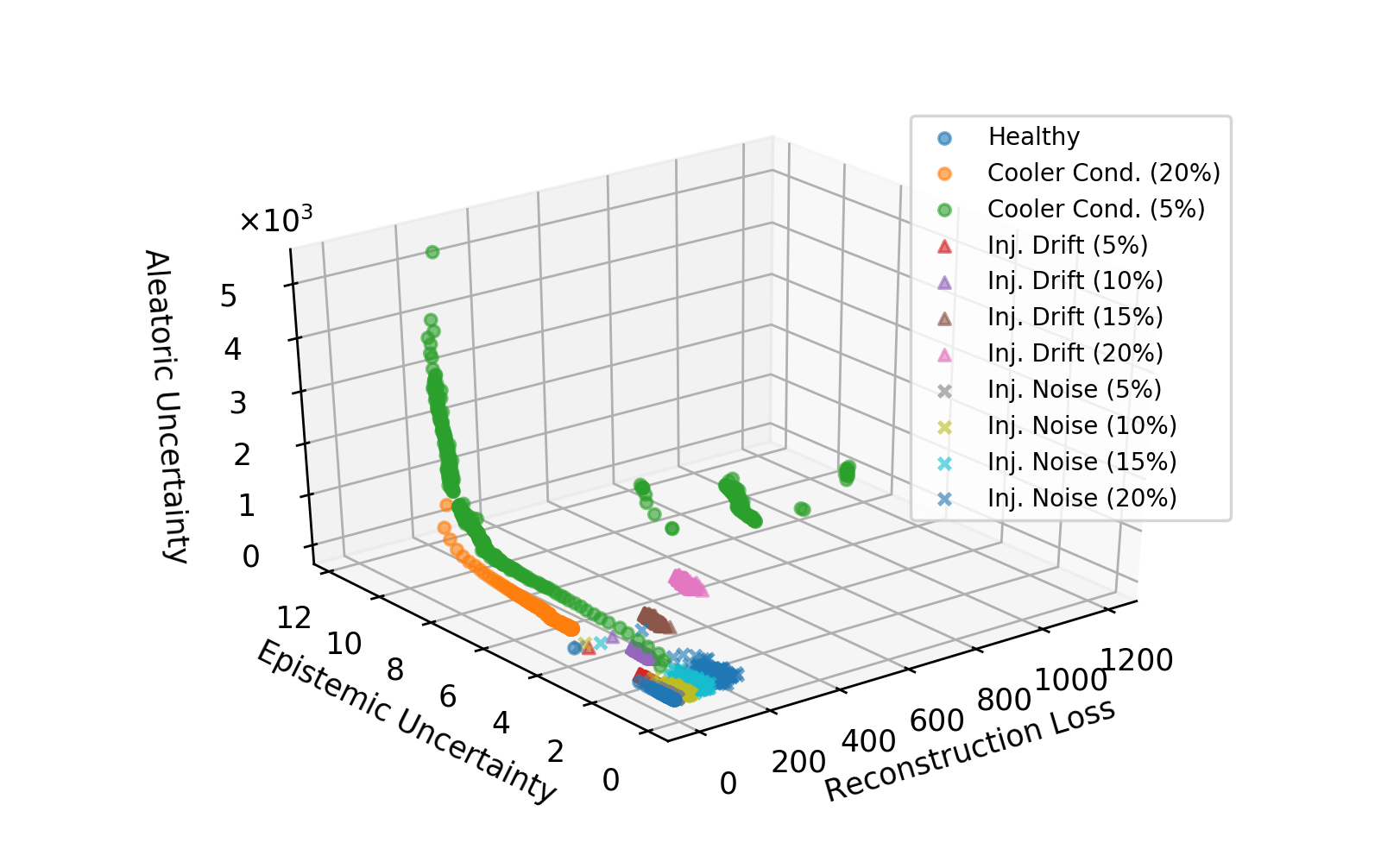}}
\caption{Scatter plot of Bayesian Autoencoder's outputs under various conditions: healthy condition, degrading cooler condition, noisy and drifting pressure sensor. This illustrates the separability of the types of drifts based on the trio: reconstruction loss, epistemic uncertainty and aleatoric uncertainty.}
\label{fig-3d-scatter-bae}
\end{figure}

By solely relying on the reconstruction loss, we are unable to distinguish real and virtual drifts. Thus, we posit that, by capturing these patterns of uncertainties, novel methods can potentially be developed to distinguish real and virtual drifts in sensor data as shown in Fig.~\ref{fig-3d-scatter-bae}. From a qualitative perspective, we note that the points form clusters which are separable. This implies we can apply a clustering algorithm (e.g k-means or hierarchical clustering) on these three metrics: reconstruction loss, epistemic uncertainty and aleatoric uncertainty of every data point to achieve unsupervised classification. To the best of our knowledge, we are the first to elicit this application within Bayesian Autoencoders.

\begin{figure}[htbp]
\centerline{\includegraphics[scale=0.65]{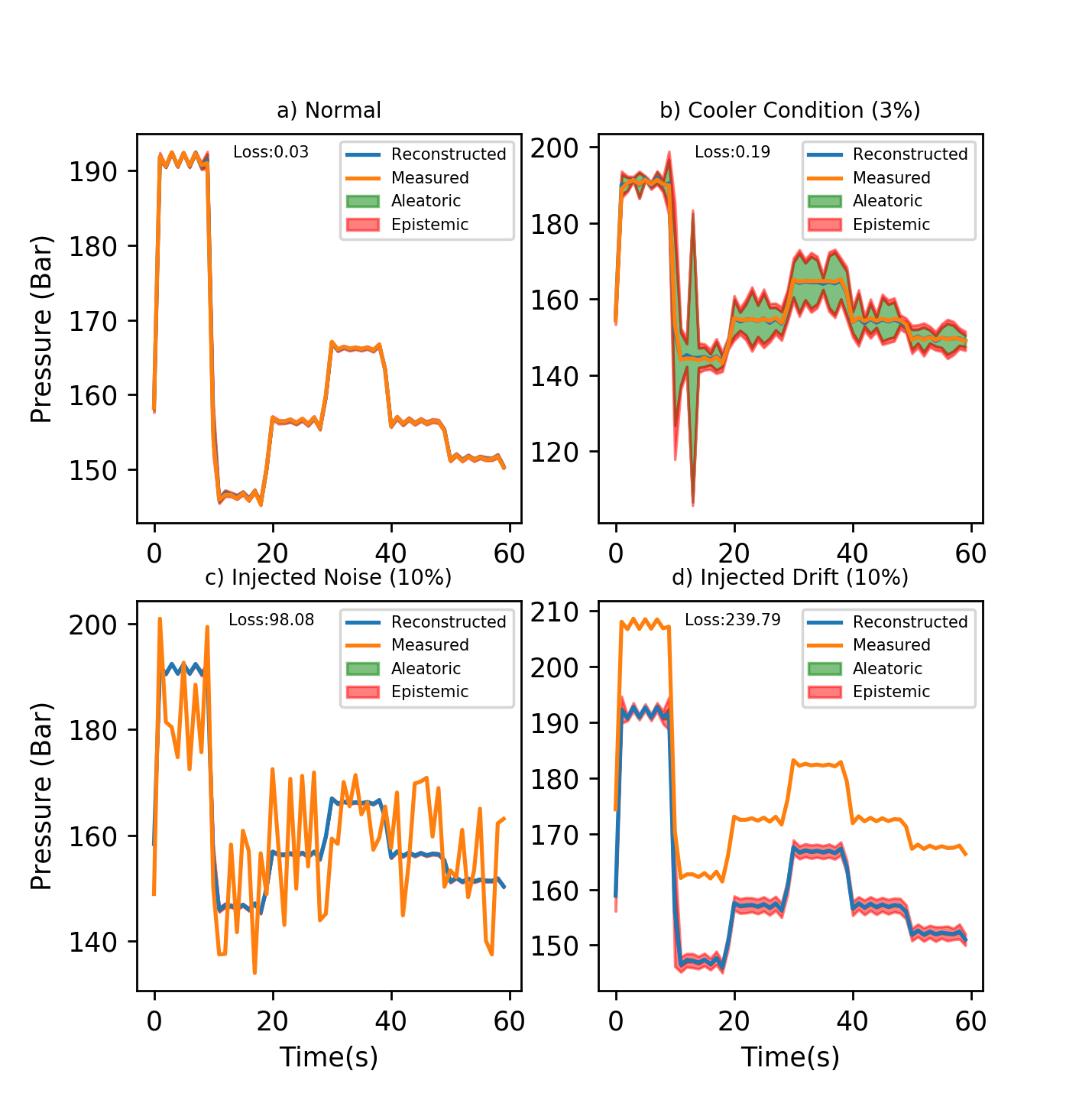}}
\caption{Measured signal and reconstructed signal of pressure sensor with epistemic and aleatoric uncertainties in a working cycle}
\label{fig-recon-sig}
\end{figure}

We have conducted further experiments (in Fig.~\ref{fig-recon-sig}) to gain more insights about the actual, reconstructed values and their uncertainties. For the nearly faulty cooler condition (Fig.~\ref{fig-recon-sig}b), the reconstruction loss shows an insignificant increase compared to the normal actual signal (Fig.~\ref{fig-recon-sig}a). However, epistemic and aleatoric uncertainties increase significantly. Despite the presence of noise and drifts (Fig.~\ref{fig-recon-sig}c \& d), we note that the Bayesian Autoencoder is able to reconstruct the shape of the normal signal. In such a case, the reconstruction loss increases rapidly; this is due to the large difference between the actual and reconstructed values. 
Meanwhile, the uncertainties do not show a significant increase in these situations. By observing the uncertainties of the reconstructed signal, operators can gain more interpretable insights into the model's predictions. Since the uncertainties are computed on a feature level, the uncertainty of every sensor on every time step can be leveraged for further decision making.

\section{Conclusion}
\label{conclusion}
Distinguishing between a real and virtual drift is of importance, especially in ML for manufacturing where the environments are highly dynamic. Our conducted experiments show that the reconstruction loss typically used in autoencoders is unable to distinguish a real drift in the environment and virtual drift due to sensor degradation. By observing the epistemic and aleatoric uncertainties, a difference is noticed in the quality of prediction in each case, which can be leveraged for distinguishing real and virtual drifts in sensors data. Since uncertainty quantification using Bayesian Autoencoders is mostly unexplored in the industrial context, this appears to be a promising field of research which deepens our understanding and trust of these deep models. We leave the detailed analysis of these observations for future studies. 

Future work will involve using a Gaussian likelihood with a full covariance matrix, instead of a diagonal only (as in our conducted experiments), which may reveal more insights in interpreting the model’s aleatoric uncertainty measures. Other than a Gaussian likelihood, the effects of using different likelihood distributions can also be explored. Moreover, we can leverage the Bayesian Autoencoder's outputs for a novel unsupervised classification method. We will also extend the experiments to study the effect of variant Bayesian Autoencoder architectures on various datasets for identifying the real and virtual drifts and their uncertainties. 

\section*{Acknowledgment}

The research presented was supported by European Metrology Programme for Innovation and Research (EMPIR) under the project Metrology for the Factory of the Future (MET4FOF), project number 17IND12 as well as the PITCH-IN (Promoting the Internet of Things via Collaborations between HEIs and Industry) project funded by Research England. We express our gratitude to Tim Pearce for his inputs. 

\bibliography{references}

\end{document}